# Multi-Modal Sarcasm Detection Based on Contrastive Attention Mechanism

Xiaoqiang Zhang, Ying Chen[*], Guangyuan Li

China Agricultural University, Beijing, China
{xqzhang,chenying, liguangy}@cau.edu.cn

**Abstract.** In the past decade, sarcasm detection has been intensively conducted in a textual scenario. With the popularization of video communication, the analysis in multi-modal scenarios has received much attention in recent years. Therefore, multi-modal sarcasm detection, which aims at detecting sarcasm in video conversations, becomes increasingly hot in both the natural language processing community and the multi-modal analysis community. In this paper, considering that sarcasm is often conveyed through incongruity between modalities (e.g., text expressing a compliment while acoustic tone indicating a grumble), we construct a Contrastive-Attention-based Sarcasm Detection (ConAttSD) model, which uses an inter-modality contrastive attention mechanism to extract several contrastive features for an utterance. A contrastive feature represents the incongruity of information between two modalities. Our experiments on MUStARD, a benchmark multi-modal sarcasm dataset, demonstrate the effectiveness of the proposed ConAttSD model.

## 1 Introduction

Sarcasm is a form of communication in which the speaker intends to communicate a contradictory situation or the opposite meaning of what is literally said. Understanding sarcasm often uses a highly complex structure of multi-modal signals [1]. For example, we employ three communicative modalities in a coordinated manner to convey our intentions: language (spoken words), vision (gestures), and audio (voice). Therefore, it is important to do multi-modal sarcasm detection that recognizes sarcasm in videos, where the three modalities are present.

According to the studies of multi-modal affective analysis (i.e., sentiment detection, emotion detection), there are dual dynamics in human communication: intra-modality dynamics and inter-modality dynamics [2]. Intra-modality dynamics refer to dynamics within each modality, and inter-modality dynamics refer to dynamics between modalities. Sarcasm detection needs the dual dynamics to find incongruity which is either from multiple modalities in an utterance or from the context of an utterance [1]. For example, there are two sarcasm cases in Fig. 1. The sarcasm in Fig. 1(a) is conveyed by an inter-

---

[*] Corresponding author

Utterance:

(a) Inter-modal incongruity

**SHELDON :**

I'm listening to you snore. I'm wondering how I'll ever sleep without it.

- **Text:** suggests neutral
- **Audio:** grumbled tone
- **Video:** gloomy face

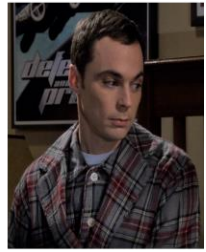

(b) Intra-modal incongruity

**LEONARD :**

Why? Because I got an **ugly**, **itchy** sweater, and my brother got a car? No, I was her **favorite**.

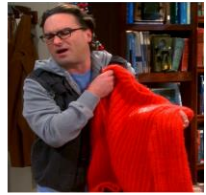

**Fig. 1.** Incongruity in sarcasm

modality incongruity, where the text indicates a compliment while the facial expressions show a disagreement. The sarcasm in Fig. 1(b) is expressed by a textual incongruity, where the sentence including "*ugly*" and "*itchy*" is indicative of negative sentiment, while the sentence including "*favorite*" gives a positive sentiment.

Although several studies [1,3] have investigated multi-modal sarcasm detection, there are remaining several unsolved research problems. One crucially problem lies in the modeling of the incongruity between modalities in an utterance. Inter-modality incongruity often plays an important role in sarcasm detection, e.g., the sarcasm in Fig. 1(a) is noticed through the contrast between the language modality and the visual modality. However, to our best knowledge, the feature extraction of the inter-modality incongruity has not yet been explored in multi-modal analysis. Therefore, more efforts are required for multi-modal sarcasm detection.

To address the issue, we propose a Contrastive-Attention-based Sarcasm Detection (ConAttSD) model, which utilizes a contrastive attention mechanism [4-5] to extract inter-modality incongruent information for multi-modal sarcasm detection. As shown in Fig. 1(a), in the feature space of the focused utterance, the features learned from the vision and audio should be similar, whereas the features learned from the text and vision should be different. Thus, we design an inter-modality contrastive attention mechanism, which produces opponent attention weights for a directed bi-modal variant (e.g., *text → audio*), and then generate a contrastive feature to represent the incongruity between the two modalities.

Overall, our main contributions can be summarized as follows:

— We design a Contrastive-Attention-based Sarcasm Detection (ConAttSD) model to detect sarcasm in video conversations.
— We propose an inter-modality contrastive attention mechanism to extract contrastive features to represent the incongruity between modalities. These contrastive features can effectively facilitate the detection of sarcasm.

– Experiments on MUStARD (a benchmark multi-modal sarcasm dataset) demonstrate the effectiveness of our ConAttSD model for multi-modal sarcasm detection.

## 2 Related Work

### 2.1 Sarcasm Detection

The studies of sarcasm detection in the textual scenario have been intensively carried out for many years. In general, text-based sarcasm detection can be divided into rule-based, statistical-machine-learning-based, and deep-learning-based.

Rule-based sarcasm detection [6-7] mainly aims to detection sentiment polarity inconsistency using different rules. Sentiment polarity inconsistency refers to two sentiments that are contradictory in their polarity (i.e., negative vs. positive). For example, the sarcasm in Fig.1(b) is expressed by sentiment polarity inconsistency. Sarcasm detection based on statistical machine learning mainly focused on feature extraction. In general, features are extracted from two perspectives [8]: the characteristics of sarcasm expressions in different-level texts (i.e., special symbols, morphology, syntax) and sentiment polarity inconsistency. In recent years, sarcasm detection based on deep learning has been explored, which uses different deep neural networks (DNNs) to extract various types of textual information. [9] used a bidirectional recurrent neural network to extract the representation of contextual information. [10] utilized multiple pre-trained models (involving emotion, sentiment, personality, etc.) to help feature extraction. [11] used two complementary adversarial learning methods to improve sarcasm detection. [12] pre-trained the BERT model to extract representations with more emotional information to help sarcasm detection.

Recently, there are concerns about sarcasm detection in multi-modal scenarios. [1] provided MUStARD, a sarcasm dataset on video conversations, and [3] assigned affective labels (sentiment and emotion) to each utterance in MUStARD. Moreover, [3] treated an utterance and its historical context as a whole by concatenating operations and then used two attention mechanisms (i.e., inter-segment inter-modal attention and intra-segment inter-modal attention) to model inter-modality dynamics. In this paper, according to the characteristics of sarcasm expressions in videos, we focus on the extracting of incongruent information between modalities in an utterance.

### 2.2 Multi-modal Affective Analysis

Multi-modal affective analysis is an important task in the multi-modal analysis community, which generalizes text-based sentiment detection or emotion detection to videos. The challenge in the multi-modal affective analysis is how to effectively model inter-modality dynamics. In general, the inter-modality modeling methods are designed for two scenarios: monologue and dialogue.

In a monologue scenario, the modeling of multi-modal interactive information mostly focuses on the exaction of sequential context information along the time axis. [2] proposed a tensor fusion network that can capture the interactive information of any modal combination. Subsequently, the network was strengthened so that the multi-modal sequence information that changes over time can be dynamically obtained [13-14]. [15] proposed a sequence-to-sequence translation model to extract multi-modal interaction information during the translation from one modality to another modality. [16] used Transformers [17] to model each modality and used its multi-head attention mechanism to capture multi-modal interaction information.

In a dialogue scenario, the affective state of a speaker is the result of the interaction of multiple factors (e.g., contextual information, previous affective state, the expression styles of the speaker). Therefore, in addition to the sequential context information, [18] used memory networks to separately model the contextual information of different speakers in dialogue. [19] used the GRU network [20] to separately model the emotions of different speakers. [21-22] used graph convolutional networks to simultaneously model the emotions of different speakers. [23] proposed two-layer Transformers, which uses Transformers to extract and modulate intra-modality information. In this paper, besides the modeling of sequential contexts and speakers in a dialogue, we mainly deal with inter-modality incongruity in an utterance, which is a specific research issue for multi-modal sarcasm.

## 3   Methodology

### 3.1   Overview

In this section, we describe our proposed ConAttSD model for multi-modal sarcasm detection. Suppose that a conversation has proceeded for $t$ turns so far with the utterance sequence $S_i = \{u_1, u_2, \ldots, u_i\}$, the $i$-th utterance $u_i$ is ready to be tested, and the other utterances are its historical context. The goal of our multi-modal sarcasm detection is to assign a binary label (1: sarcasm; 0: no sarcasm) to $u_i$ conditioned on $u_i$ and its historical context.

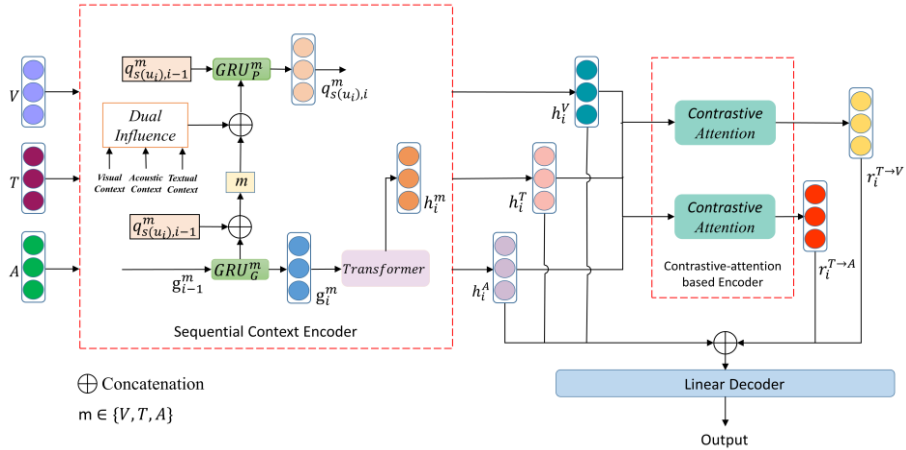

**Fig. 2.** The overview of our multi-modal sarcasm detection

Specifically, as illustrated in Fig. 2, our ConAttSD model comprises two encoders (sequential context encoder and contrastive-attention-based encoder) and one decoder. The sequential context encoder dynamically captures intra-modality influence transmitted along with the conversation, and the contrastive-attention-based encoder extracts incongruent information between modalities in $u_i$ by an inter-modality contrastive attention mechanism. Then, a linear decoder assigns a sarcasm label to $u_i$ according to its representation.

### 3.2 Utterance Representation

Each utterance is represented by three vectors: a textual feature *T* for linguistic content (text), an acoustic feature *A* for acoustic characteristics (audio), and a visual feature *V* for visual information (vision).

**Textual Feature Extraction**

Textual feature *T* is generated by a pre-trained BERT model [24] and the dimension is 768 ($d_t = 768$).

**Acoustic Feature Extraction**

The library Librosa [25] is used to extract acoustic features, including Mel-frequency cepstral coefficients (MFCCs), pitch tracking and voiced/unvoiced segmenting features, peak slope parameters, and maxima dispersion quotients. The average of the acoustic features for the focused utterance is used as acoustic feature *A*, and the dimension is 298 ($d_a = 298$).

**Visual Feature Extraction**

A pre-trained Resnet-152 [26] is used to extract a visual feature for a visual frame. The average of visual features of all frames in the focused utterance is used as visual feature *V*, and the dimension is 2048 ($d_v = 2048$).

### 3.3 Sequential Context Encoder

Since conversations are highly sequential in nature and contextual information flows along a sequence of utterances, we propose a sequential context encoder to model the inter-modality influence between these utterances. As shown in Fig. 2, there are two sub-encoders used in the sequential context encoder: a GRU-based encoder which uses GRU [20] to extract sequential context information, and a Transformer-based encoder which applies Transformers [17] to the vector output from the GRU-based encoder.

In the GRU-based encoder, we adopt the inter-modal influence modeling proposed by [27] for multi-modal sentiment detection. For utterance $u_i$, global state $g_i^m$ and speaker state $q_{s(u_i),i}^m$ interact together to represent both utterance $u_i$ and speaker $s(u_i)$ (i.e., the speaker of the utterance $u_i$), respectively. Notice that a global state $g_i^m$ is actually a kind of sequential context information for utterance $u_i$. Formally, both the global state and the speaker state are iteratively updated as follows.

$$g_i^m = GRU_G^m((u_i^m \oplus q_{s(u_i),i-1}^m), g_{i-1}^m) \quad (1)$$

$$q_{s(u_i),i}^m = GRU_P^m((u_i^m \oplus c_i^m), q_{s(u_i),i-1}^m) \quad (2)$$

where $m \in \{T, A, V\}$ is the modality, $g_i^m$ denotes the global state representation of $i$-th turn utterance for the modality $m$, $q_{s(u_i),i}^m$ denotes the speaker state representation at $i$-th turn utterance for the modality $m$, $c_i^m$ is the context representation of $i$-th utterance using a dual influence network, which includes both intra-modal and inter-modal information.

In the Transformer-based encoder, we extract more effective sequential context information using Transformers, which have shown superior performance in capturing long-range dependency than RNN models. A Transformer is composed of a stack of $B$ identical blocks, and each block has two sub-layers (including a multi-head self-attention mechanism and a Multi-Layer Perceptron) with a residual connection, as shown in Eq.3. In this paper, we use a Transformer to capture the dependency inside the global state $g_i^m$ ($m \in \{T, A, V\}$) and output a sequential context vector $h_i^m$ for utterance $u_i$, as illuminated in Fig. 2.

$$y = LayerNorm(x + Sublayer(x)) \tag{3}$$

### 3.4 Contrastive-attention-based Encoder

To extract the incongruent information between multiple modalities for sarcasm detection, we propose an inter-modality contrastive attention mechanism that applies the contrastive attention mechanism to the three sequential context vectors ($h_i^T$, $h_i^A$ and $h_i^V$) outputted from the sequential context encoder.

**Contrastive Attention.**

The contrastive attention mechanism, which attempts to capture irrelevant or less relevant parts between two vectors, was proposed by [4] for person re-identification (a computer vision task), and then used for text summarization (an NLP task) by [28]. In fact, the contrastive attention mechanism is transformed from the self-attention mechanism used in Transformers.

Specifically, given three input vectors Q, K and V, the self-attention mechanism is defined by Eq. 4, and the contrastive attention mechanism is defined by Eq. 5-7. First, the attention weights $a_c$ is calculated by Eq. 5. Then, the opponent attention weights $a_o$ is obtained through the opponent function applied on $a_c$ followed by the softmax function, as shown in Eq. 6. Compared to the conventional attention weights $a_c$ which capture the most relevant part between $Q$ and $K$, the opponent attention weights $a_o$ focuses on their irrelevant parts. Lastly, a contrastive vector is generated by Eq. 7, which is the weighted sum of elements of $V$ and the weights are the opponent attention weights.

$$y = softmax(\frac{QK^T}{\sqrt{d_k}})V \tag{4}$$

$$a_c = softmax(\frac{QK^T}{\sqrt{d_k}}) \tag{5}$$

$$a_o = softmax(1 - a_c) \tag{6}$$

$$r = a_o V \tag{7}$$

7where $Q$, $K$, $V$ are queries, keys and values, respectively, and $d_k$ is the dimension of $K$.

**Inter-modality Contrastive Attention.**

In an inter-modality contrastive attention, a directed bi-modal variant (e.g., $T \rightarrow A$: $Q = h_i^T$ and $K = h_i^A$) is used as its inputs, and an inter-modality contrastive vector (e.g., $r_i^{T \rightarrow A}$) is generated for utterance $u_i$. First, opponent attention weights are learned by Eq. 6. Then, the opponent attention weights are applied to a modality (e.g., $V = h_i^A$) to produce the inter-modality contrastive vector by Eq. 7.

As shown in Fig. 2, our contrastive-attention-based encoder takes the textual modality as the anchor modality, and generates two directed bi-modal variants (i.e., $T \rightarrow A$ and $T \rightarrow V$) as input for two inter-modality contrastive attentions, respectively. Then, two inter-modal contrastive vectors (i.e., $r_i^{T \rightarrow A}$ and $r_i^{T \rightarrow V}$) are generated, where an inter-modal contrastive vector (e.g., $r_i^{T \rightarrow A}$) represents the incongruity in its corresponding input bi-modal variant (e.g., $T \rightarrow A$). Thus, through the two inter-modality contrastive attentions, text can be effectively contrasted with information from audio and vision, respectively.

### 3.5 Linear Decoder

As illuminated in Fig. 2, the three sequential context vectors from the sequential context encoder and the two inter-modal contrastive vectors from the contrastive-attention-based encoder are concatenated as Eq. 8. Then, the final vector $\beta_i$ is feed to a softmax classifier to obtain the sarcasm label of utterance $u_i$.

$$\beta_i = [h_i^T, h_i^A, h_i^V, r_i^{T \rightarrow A}, r_i^{T \rightarrow V}] \tag{8}$$

## 4 Experiment

### 4.1 Setup

**Datasets.**

In our experiment, we use a benchmark Multimodal Sarcasm Detection Dataset (MUStARD) provided by [1] for multi-modal sarcasm detection. The dataset was collected from 4 popular TV Series: Friends, the Big Bang Theory, the Golden Girls, and Sarcasmaholics Anonymous. There are totally 690 samples (i.e., conversations) with an even number of sarcastic and non-sarcastic samples, and the utterances in each sample consist of three modalities: vision, audio, and text.

Moreover, there are two experimental setups to use MUStARD for the sarcasm detection[1]: speaker-dependent and speaker-independent. Compared to the speaker-dependent scenario, the speaker-independent setup is more challenging because it prevents the detection model using registered speaker's specific information and requires the model with a higher degree of generalization. Since it is often a case that a unregistered speaker appears in a video conversation, we work on the sarcasm detection with the speaker-independent setup in this paper. Specifically, the speaker-independent dataset split is used in our experiment, where videos from The Big Bang Theory, The Golden Girls, and Sarcasmaholics Anonymous are served as the training set and videos from Friends are used as the testing set. Moreover, the performances are evaluated by three metrics: precision (P), recall (R), and F1-score (F1).

**Baselines.**

We compare our proposed approach with the following two models using different encoders.

- Two-attention-based encoder: This is a multi-task learning system developed by [3] to detect simultaneously detect sarcasm, sentiments, and emotions. For each modality, the focused utterance and its historical context utterances are concatenated as an input utterance. Then, two attention mechanisms (i.e., inter-segment inter-modal attention and intra-segment inter-modal Attention) are applied to model inter-modality dynamics for the input utterance. Lastly, a multi-task learning framework is used based on inter-modality interactive information.
- GRU-based encoder: This is a pure sarcasm detection system, which concatenates three global states from our GRU-based encoder (i.e., $g_i^T$, $g_i^A$ and $g_i^V$) to detect sarcasm in the focused utterance.

**Implementation Details.**

While training, we use the Adam optimizer [29] to update all hyper-parameters. Each training batch contains 64 conversations, and the learning rate is set to 0.0001. For GRU, to reduce over-fitting, dropout [30] is applied, and it is set 0.5. The size of the hidden representations is 150. For Transformers, the number of blocks $B$ is set to 3, and the number of heads is 6.

### 4.2 Results and Analysis

**Model Comparison.**

We first compare our ConAttSD model with the two baseline models, and list the performances in Table 1.

Table 1. Sarcasm detection results of different models.

|  | P | R | F1 |
|---|---|---|---|
| Two-attention-based encoder [3] | 71.51 | 71.35 | 70.46 |
| GRU-based encoder | 71.69 | 70.90 | 70.82 |
| Sequential Context Encoder (GRU+Transformer) | 72.32 | 72.32 | 72.26 |
| ConAttSD (GRU+Transformer+Contrastive Attention) | **74.46** | **74.01** | **73.97** |

**Table 2.** Sarcasm detection results of the sequential context encoder using different modalities.

|             |       | P     | R     | F1    |
|-------------|-------|-------|-------|-------|
| Uni-modal   | T     | 53.38 | 53.39 | 53.39 |
|             | A     | 67.99 | 67.23 | 66.52 |
|             | V     | 71.42 | 71.19 | 71.19 |
| Multi-modal | T+A   | 60.38 | 60.45 | 60.35 |
|             | T+V   | 70.42 | 70.34 | 70.35 |
|             | A+V   | 73.17 | 72.03 | 71.89 |
|             | T+ A+V | **72.32** | **72.32** | **72.26** |

**Table 3.** Sarcasm detection results of the contrastive-attention-based encoder using different bi-modal variants.

|                                   | P     | R     | F1    |
|-----------------------------------|-------|-------|-------|
| $T \to A$                         | 71.24 | 71.19 | 71.07 |
| $A \to T$                         | 71.51 | 70.90 | 70.85 |
| $T \to V$                         | 70.29 | 70.06 | 70.06 |
| $V \to T$                         | 72.19 | 72.97 | 72.58 |
| $A \to V$                         | 70.66 | 70.34 | 70.33 |
| $V \to A$                         | 70.91 | 70.62 | 70.62 |
| Optimal: $T \to A + T \to V$      | **74.46** | **74.01** | **73.97** |

In Table 1, ConAttSD significantly outperforms the two baseline systems. E.g., compared to the best baseline system (i.e., the GRU-based encoder), the F1 score of ConAttSD rises by 3.51%. Specifically, the F1 score rises by 1.44% (from 70.82% to 72.26%) through incorporating the Transformer-based encoder and furthermore increases 1.71% (from 72.26% to 73.97%) by adding the contrastive-attention-based encoder. This indicates that our Transformer-based encoder and contrastive-attention-based encoder can effectively extract sequential context information and the inter-modality incongruent information.

Moreover, from Table 1, we observe that although the baseline model, the two-attention-based encoder, adopts a multi-task learning framework to use richer label information (sentiment, emotion, and sarcasm), its performance is still not comparable to the one of ConAttSD whose input is only sarcasm labels. This indicates that how to effectively combine affective information and sarcasm for multi-modal analysis needs more investigation.

**Modality Comparison.**
To further explore the effects of the three modalities for our ConAttSD model, we perform an in-depth analysis of the sequential context encoder and the contrastive-attention-based encoder with different modalities, respectively.

First, we evaluate our sequential context encoder with all possible inputs: uni-modal variants (i.e., *T*, *A*, and *V*), bi-modal variants (i.e., *T+A*, *T+V*, and *A+V*), and a tri-modal variant (i.e., *T+A +V*), the performances are shown in Table 2. In Table 2, the model

44410

with the visual modality achieves the best performance among the unimodal variants. Furthermore, the addition of acoustic modality (i.e., $A+V$) slightly improves the unimodal baseline (from 71.19% to 71.89% in F1 scores). Finally, the tri-modal variant achieves the best performance (72.26% in F1 scores).

Then, based on the optimal sequential context encoder whose input is $T+A+V$, we evaluate our contrastive-attention-based encoder with all possible inputs. Notice that the inter-modality contrastive attention requires exactly two modalities, and any directed bi-modal variant (e.g., $A \rightarrow T$, $A \rightarrow V$) can serve as input to the inter-modality contrastive attention. The performances are listed in Table 3. In Table 3, the model input with either $V \rightarrow T$ or $T \rightarrow A$ achieves good performances among these directed bi-modal variants. E.g., the F1 score is 72.58% for $V \rightarrow T$ and 71.07% for $T \rightarrow A$. This indicates that texts that look seemingly straightforward is noticed to contain sarcasm only when vocal tonality and facial expressions are taken into account. After searching all possible combinations of the directed bi-modal variants, we find that the model with the two directed bi-modal variants (i.e., $T \rightarrow V$ and $T \rightarrow A$) achieves the best performance (73.97% in F1 scores).

## 5  Conclusion

In this paper, we propose a novel Contrastive-Attention-based Sarcasm Detection (ConAttSD) model for multi-modal sarcasm detection. Experimental results indicate the capability of our ConAttSD in capturing inter-modal incongruent information by inter-modality contrastive attention. In the future, we would like to investigate the combination of sarcasm and affective information for multi-modal analysis.